\documentclass{article}

\usepackage[final, nonatbib]{nips_2017}

\usepackage[utf8]{inputenc} 
\usepackage[T1]{fontenc}    
\usepackage{hyperref}       
\usepackage{url}            
\usepackage{booktabs}       
\usepackage{amsfonts}       
\usepackage{nicefrac}       
\usepackage{microtype}      
\usepackage{amsmath, graphicx}
\usepackage{caption}
\usepackage{tabularx}

\def\bibfont{\footnotesize}

\title{Distributed Stratified Locality Sensitive Hashing for Critical Event Prediction in the Cloud}

\author{
  Alessandro De Palma, Erik Hemberg, Una-May O’Reilly \\
  Computer Science and Artificial Intelligence Laboratory \\
  Massachusetts Institute of Technology\\
  \texttt{adepalma@mit.edu, \{hembergerik,unamay\}@csail.mit.edu} \\
}

\begin{document}

\maketitle

\begin{abstract}
	The availability of massive healthcare data repositories calls for efficient tools for data-driven medicine.
	We introduce a distributed system for Stratified Locality Sensitive Hashing to perform fast similarity-based prediction on large medical waveform datasets. Our implementation, for an ICU use case, prioritizes latency over throughput and is targeted at a cloud
	environment.
	We demonstrate our system on Acute Hypotensive Episode prediction from Arterial Blood Pressure waveforms. On a dataset of $1.37$ million points, we show scaling up to $40$ processors and a $21\times$ speedup in number of comparisons to parallel exhaustive search at the price of a $10\%$ Matthews correlation coefficient (MCC) loss. Furthermore, if additional MCC loss can be tolerated, our system achieves speedups up to two orders of magnitude. 
\end{abstract}

\section{Introduction}

Physiological time-series data from medical sensors is constantly increasing: in order to utilize this data to inform clinical decision support, scalable yet interpretable algorithms are needed. K-Nearest-Neighbors (KNN) methods have long been the golden standard for time series prediction \cite{Chen13, Xi06} and still offer greater interpretability than the state-of-the-art (e.g., CNN \cite{Wang16}). 
Exact KNN methods have a linear prediction time in the size of the dataset for high-dimensional points. Therefore, previous work analyzed the effectiveness of fast approximate nearest neighbor algorithms for prediction on time-series and, in particular, of locality sensitive hashing (LSH) on arterial blood pressure (ABP) datasets \cite{SLSHretrieval}, \cite{LSHprediction}. 
LSH incurs superlinear memory costs: to address this limitation and to benefit from the parallelism of modern computer architectures, distributed and parallel LSH algorithms have been presented \cite{Bahmani}, \cite{Haghani}, \cite{Sundaram}. Moreover, a distributed LSH implementation with limited functionalities is available on Spark's MLlib \cite{Spark}.
Here, we introduce our own distributed LSH system, relying on Stratified Locality Sensitive Hashing (SLSH), a multi-metric implementation of LSH with demonstrated effectiveness on medical time-series \cite{SLSHretrieval}.
Our design is targeted at the Intensive Care Unit (ICU) setting, in which the speed for a single query matters more than high-volume query processing. Furthermore, our implementation is made for the cloud environment for ease of access with respect to a private compute cluster.
We showcase the system's performance on Acute Hypotensive Episodes (AHE) prediction on datasets in the order of one million points from the MIMIC-III waveform database \cite{MIMICIII}, using up to $200$ times more data than the previous LSH ABP analyses. We demonstrate scaling up to $40$ processors and show that previous LSH results in the domain still hold on larger datasets.
In the following, we first review SLSH ($\S 2$), then describe our distributed SLSH system ($\S 3$) and finally present our experimental results ($\S 4$).

\section{Background: Locality Sensitive Hashing}
\label{background}
Standard Locality Sensitive Hashing is based on building multiple
 hash tables where collisions are maximised for similar data points~ \cite{IM98}. Within such tables, a point and its nearest neighbors will likely lie in the same hash buckets, allowing for a sublinear time algorithm at the price of superlinear memory requirements.
LSH relies on hash functions that are $(r, cr, p_1, p_2)$-sensitive \cite{IM98} for a given distance metric $\delta(\cdot,\cdot)$, i.e., hash functions belonging to a family $\mathcal{H}={h:M\rightarrow U}$ such that, for any two points $x,y \in M$, a constant $c\geq1$, and two probabilities $p_1 > p_2$: if $\delta(x, y) \leq r$, then $P_{\mathcal{H}}[h(x)=h(y)] \geq p_1$ and if $\delta(x, y) \geq cr$, then $P_{\mathcal{H}}[h(x)=h(y)] \leq p_2$. We use two such hash families: the bit-sampling family when $\delta$ is the $l_1$ norm \cite{GIM99} and the Random Projection family for the Cosine norm \cite{Charikar}.
In order to decrease the number of collisions, a new hash family $\mathcal{H^\prime}={h:M\rightarrow U^m}$ is made from $m$ independent hash functions in $\mathcal{H}$.
Retrieval recall, i.e., the number of neighbors matching those retrieved by an exhaustive search, is then increased by using $L$ independent tables indexed by instances of $\mathcal{H^\prime}$.
A set of \emph{candidates} for a query is obtained by the union of the datapoints which collide with the query in the $L$ tables. A linear search is then performed on the candidates (rather than on the entire dataset) to answer the query.
The performance of LSH is influenced by $m$ and $L$ \cite{Dong2008}.

\textbf{Stratified Locality Sensitive Hashing.}
Standard LSH compares two datapoints according to one metric. Often, this does not suffice for complex domains such as medical time series and results in a gap between semantic and metric spaces \cite{SLSHretrieval}.
Moreover, for large datasets, the linear search over the candidates is the bottleneck for LSH \cite{LSHcomparison}.
Stratified Locality Sensitive Hashing (SLSH)\cite{SLSHretrieval} addresses these drawbacks and demonstrated both speedups and a higher prediction accuracy in the ABP domain. 
Let us denote by $n$ the number of points in the dataset. SLSH uses the points in the most populous buckets, i.e., those which contain more than $\alpha n$ candidates, as the population for a further (inner) layer of LSH, employing a different metric. This not only reduces the number of candidates for a query, but also incorporates another notion of similarity in their selection \cite{SLSHretrieval}.
For this application, we employed $l_1$ norm for the outer layer and cosine similarity for the inner layer. We denote the parameters for the outer LSH layer with the subscript $out$ (e.g., $m_{out}$) and those for the inner layer with $in$ (e.g., $m_{in}$).

\section{Distributed SLSH}
\label{system}


\begin{figure}[b]
	\begin{minipage}{.45\textwidth}
		\centering
		\includegraphics[height=110pt]{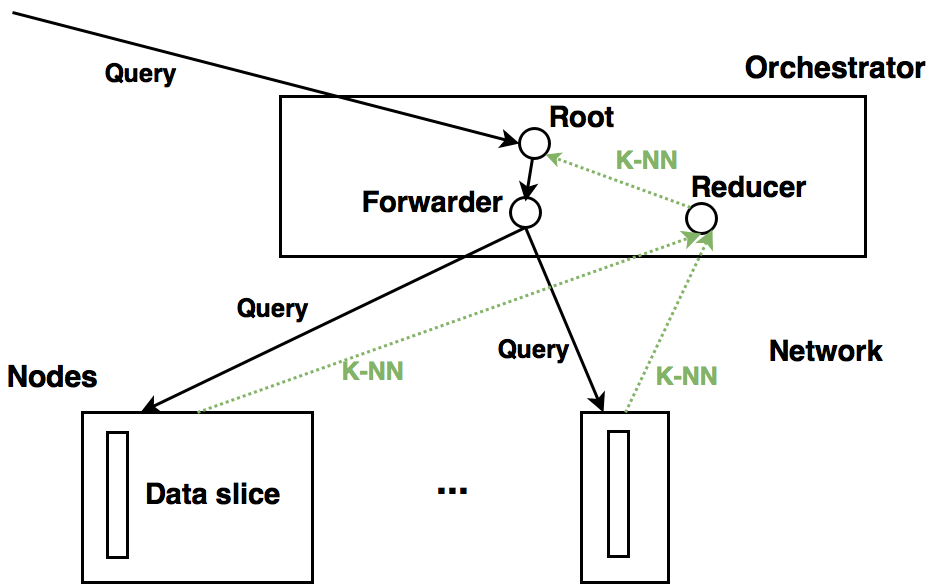}
		\caption{Multi-node distributed system architecture.}
		\label{fig:distributed}
	\end{minipage} \hspace{5pt}
	\begin{minipage}{.45\textwidth}
		\centering
		\includegraphics[height=110pt]{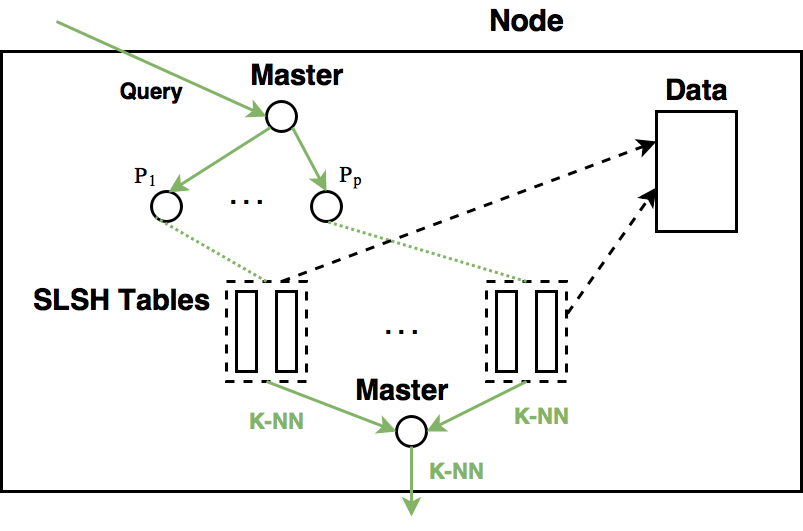}
		\caption{Intra-node architecture; cores are denoted by $P_i$.}
		\label{fig:intranode}
	\end{minipage}
\end{figure}

We now describe the design of our distributed SLSH system (DSLSH), which is publicly available at \url{https://github.com/Distributed-SLSH}.
Figure \ref{fig:distributed} shows the architecture of the distributed system, which is composed of $\nu$ identical SLSH nodes of $p$ cores and an \textit{Orchestrator}, coordinating the execution. Each of the nodes executes a parallel SLSH algorithm on a subset of the dataset of $O(n/\nu)$ points, as in Sundaram et al. \cite{Sundaram}.
The \textit{Orchestrator} uses three processes, called \textit{Root}, \textit{Forwarder} and \textit{Reducer}. 
The \textit{Root} coordinates query resolution and manages the table construction on the nodes.
\textit{Forwarder} and \textit{Reducer} respectively forward the queries to the nodes and process their local outputs. During table construction, the \textit{Root} assigns each node its share of the dataset and broadcasts the $m_{out}$ $L_{out}$ outer hash functions, as the same hash family instances need to be used. Using these hashes, each node then builds the tables in parallel with its dataset slice. 
When a query arrives, it is sent from the \textit{Root} to the \textit{Forwarder}, which broadcasts it to the nodes. Then, each node performs the query resolution in parallel and sends its local approximate K-NN to the \textit{Orchestrator}. These local outputs are gathered at the \textit{Reducer}, which yields the global K-NN set by keeping the K closest candidates to the query (reduction operation). The global K-NN set is used to output the final prediction result at the \textit{Root}.

\textbf{Intra-node Parallelism.}
We parallelize on the multiple tables of the outer LSH layer, assuming that $p < L_{out}$. This usually holds in practice, as a rather large $L_{out}$ is needed to retrieve a good approximation of the nearest neighbor set. An alternative approach to intranode parallelism is to batch the queries and rely on the fact that queries distant from each other will not hash to the same buckets \cite{Sundaram}. Such a design would be ill-suited for our ICU use-case, as we expect a low number of queries per second.
Figure \ref{fig:intranode} outlines the architecture within a distributed SLSH node.
The dataset is stored in shared memory and each processor $P_i$ has $O(L_{out}/p)$ outer tables, whose buckets contain pointers to the shared memory.
These tables are constructed entirely in parallel, as each table employs independent instances of $\mathcal{H^\prime}$. Therefore, there is no overlap in the computations for any pair of hashes.
Inner LSH tables are then built sequentially where the population is larger than $n \alpha$.
When a query from the \textit{Orchestrator} is received, the arbitrary \textit{Master} process broadcasts it to the other cores. Query resolution (possibly resorting to the second LSH layer) is then performed locally by each core on the share of tables it owns, yielding a partial K-NN set. These partial results are gathered at the \textit{Master}, which computes the final output through a reduction and sends it to the \textit{Orchestrator}.


\section{Experiments and Results}

\begin{table}[t]
	\small
	\centering
	\caption{Employed ABP datasets for AHE prediction. $l$ and $c$ denote lag and condition window lengths, $l/d$ is the length of the lag subwindows that the samples of the series represent. The dataset for the previous LSH AHE analyses is provided for reference.}
	\begin{tabular}{llllcc}
		\toprule
		Name     & $l$ & $l/d$ & $c$ & $n$ points (scale-up to \cite{SLSHretrieval}, \cite{LSHprediction}) &  $\%\overline{\text{AHE}}$ \\
		\midrule
		\texttt{AHE-30l-30c} & $30$ min. & $1$ min. & $30$ min. & $8.037 \times 10^5$ ($124$) & $98.45\%$ \\
		\texttt{AHE-5l-5c}   & $5$ min. & $10$ s & $5$ min. & $1.373 \times 10^6$ ($212$) & $96.04\%$  \\
		Kim et al. \cite{SLSHretrieval}, \cite{LSHprediction}   & $300$ min. & $1$ min. & $30$ min. & $6.467 \times 10^3$ ($1$) & $92.06\%$  \\
		\bottomrule
	\end{tabular}
	\label{table}
\end{table}

\begin{figure}[b]
	\begin{minipage}{.45\textwidth}
		\centering
		\includegraphics[height=100pt]{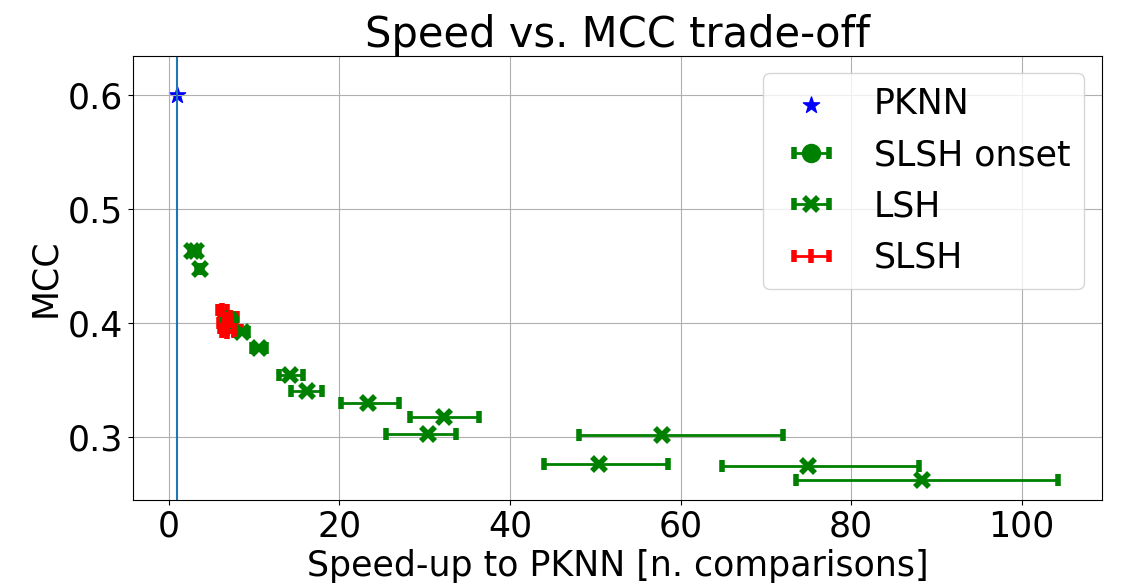}
		\caption{Speedup and MCC loss to PKNN on dataset \texttt{AHE-30l-30c}, with $p=8$, $\nu=2$. MCC $\in [-1,1]$, the higher the better. }
		\label{fig:speedmcc}
	\end{minipage} \hspace{15pt}
	\begin{minipage}{.45\textwidth}
		\centering
		\includegraphics[height=100pt]{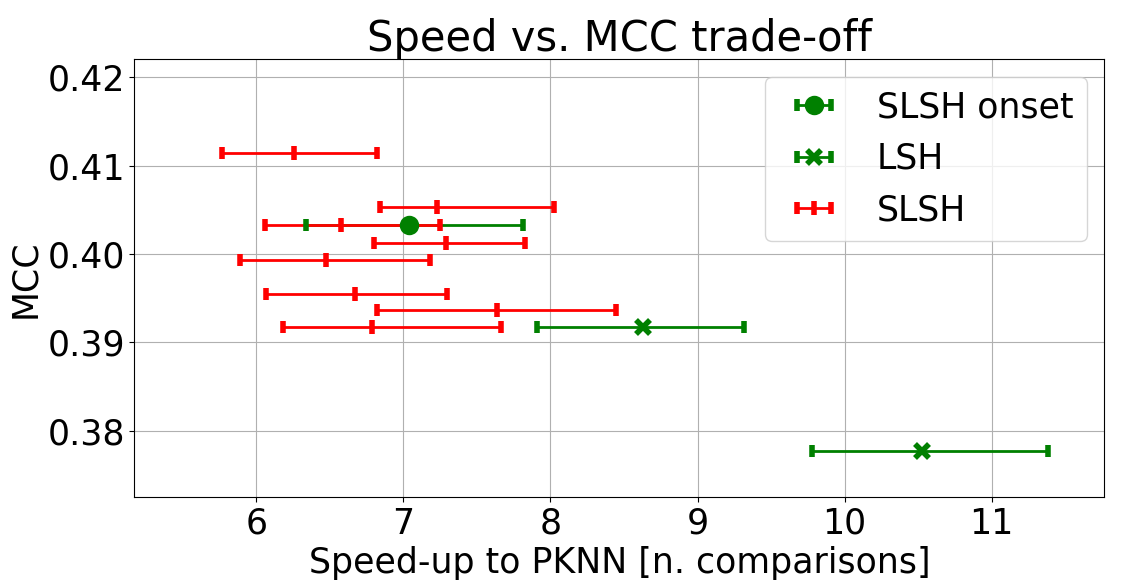}
		\caption{Zoom-in on Figure \ref{fig:speedmcc}. $p=8$, $\nu=2$. SLSH onset denotes the outer LSH configuration on which the inner layer is applied.}
		\label{fig:speedmcczoom}
	\end{minipage}
\end{figure}

We present experimental results on the prediction of acute hypotensive episodes on arterial blood pressure data extracted from the MIMIC-III ICU waveform database \cite{MIMICIII}, freely accessible from PhysioNet \cite{PhysioNet}.
We first extend previous LSH ABP analyses (\cite{LSHprediction}, \cite{SLSHretrieval}) on larger datasets and show various prediction quality trade-offs ($\S$\ref{tradeoff}).  Then, for a fixed trade-off (as parallelism does not influence the prediction output), we test whether our DSLSH system properly employs the available compute resources with a scalability analysis ($\S$\ref{scalability}).

Our datasets are composed of time series spanning a time window of length $l$ (lag window) with $d=30$ subwindows, each of length $l/d$. The average Mean Arterial Pressures (MAP) of the valid heart beats within each subwindow form the $d$ samples of the series. Beat validity is assessed by checking whether each beat respects a set of properties~\cite{BeatDB}. These time-series are used for similarity search and are labeled positive or negative depending on whether an AHE occured in a time interval of length $c$ immediately following the lag window (condition window).
We define AHE as a $c$-minute interval in which at least $90\%$ of the per-beat MAP values are below $60$ mmHg.
Depending on the size of the lag and condition windows, we obtain datasets of different sizes by applying a rolling window algorithm on all the available ABP waveforms. The algorithm moves the window forward by $10\%$ of the total window size ($l + c$) in case of no AHEs and immediately after the previous window if an AHE is present \cite{BeatDB}.
Table~\ref{table} describes the employed datasets and compares them to previous work. As a result of the rolling window algorithm, waveforms without AHE ($\%\overline{\text{AHE}}$) dominate our datasets.


\subsection{Speed vs. Matthew's Correlation Coefficient trade-off}
\label{tradeoff}

We investigate speed vs. prediction quality as a function of the SLSH parameters, using weighted voting with $K=10$ nearest neighbors for prediction. The goal is to extend the results in \cite{SLSHretrieval} on larger datasets. Therefore, we focus on \texttt{AHE-30l-30c}, where the condition window length is the same as in \cite{SLSHretrieval} but with a factor $124$ more data, a higher class imbalance and a lag window $10$ times smaller.
We resort to Matthew's Correlation Coefficient (MCC) for prediction quality, a robust measure in cases of severe class imbalance such as ours \cite{Powers2011EvaluationFP}. For speed, we measure the maximum number of comparisons (distance computations) across all processors, the bottleneck for large datasets \cite{LSHcomparison}. We report speedup with respect to a data-parallel $l_1$ exhaustive search for K-NN (PKNN). Data-parallel exhaustive search assigns equal shares of the points to all the processors in all the nodes, resulting in $\frac{n}{p\nu}$ comparisons per processor.
Figures \ref{fig:speedmcc} and \ref{fig:speedmcczoom} show median values and $95\%$ Confidence Intervals for the speedup of each parameter combination, on an out-of-sample test set of $2000$ queries with $p=8$, $\nu=2$. We first employ only the outer LSH layer (LSH) with $m_{out} \in \{100,\ 125,\ 150,\ 175,\ 200\}$, $L_{out} \in \{72,\ 96,\ 120\}$. Then, for the point having the best speedup and at most $0.2$ ($10\%$) loss in MCC (SLSH onset: $m_{out}=125$, $L_{out}=120$), we repeat the experiments with an inner layer and $m_{in} \in \{40,\ 65,\ 90,\ 115\}$, $L_{in} \in \{20,\ 60\}$, $\alpha=0.005$ (SLSH). As expected from Section \ref{background}, increasing $m$ reduces the MCC and increases the speedup, while increasing $L$ has the opposite effect. We expect with tuning that the inner layer would increase speedup and could simultaneously increase MCC (due to the use of a second distance metric) compared to its onset. We point out that the outer layer yields a wide range of trade-offs (Figure \ref{fig:speedmcc}): depending on the given ICU use-case, the appropriate SLSH onset should be chosen according to the MCC loss a clinician can tolerate.

\subsection{Scalability Analysis}
\label{scalability}
 
This section examines the behavior of DSLSH on two datasets of different size, as the overall number of processors, $p \nu$, increases (strong scaling). 
We focus on speed vs. MCC trade-offs tolerating a $10$-$11\%$ MCC loss and use PKNN as a baseline.
Tables \ref{fig:strong3030} and \ref{fig:strong55} show results for $p=8$, $\nu \in \{1,\dots, 5\}$ and report the median (and its $95\%$ CI) of the maximum number of comparisons across the processors on $2000$ queries.
We see that DSLSH has almost perfect scaling when adding nodes for both \texttt{AHE-5l-5c} and \texttt{AHE-30l-30c}. 
Moreover, as expected from the sublinear dependence on dataset size of LSH, the speedup of DSLSH to PKNN increases from \texttt{AHE-30l-30c} to \texttt{AHE-5l-5c}. Therefore, we can expect such a speedup to further increase on larger datasets, at no additional MCC loss.

\begin{table}[t]
	\centering
	\begin{minipage}{.45\textwidth}
		\centering
		\caption*{\small $n=801725$, median \#comparisons ($\times 10^3$)}
		\resizebox{\columnwidth}{!}{%
			\begin{tabular}{rrcrc}
				\toprule
				$\nu p$     & DSLSH ($S_8$) & DSLSH CI & PKNN & PKNN/DSLSH \\
				\midrule
				8  & 9.58 (1.00) & [8.83, 10.57] & 100.23 & 10.46  \\
				16 & 5.60 (1.71) & [4.90, 6.39]  &  50.11 &  8.94 \\
				24 & 3.36 (2.85) & [2.99, 3.79]  &  33.40 &  9.93 \\
				32 & 2.47 (3.88) & [2.26, 2.71]  &  25.05 & 10.14  \\
				40 & 2.32 (4.12) & [2.08, 2.56]  &  20.04 &  8.63 \\
				\bottomrule
			\end{tabular}
		}
		\caption{\small Strong scaling on \texttt{AHE-30l-30c}, tolerated MCC loss: $11\%$. $S_8$ denotes speedup to single-node ($\nu p = 8$).}
		\label{fig:strong3030}
	\end{minipage} \hspace{15pt}
	\begin{minipage}{.442\textwidth}
		\centering
		\caption*{\small $n=1371479$, median \#comparisons ($\times 10^3$)}
		\resizebox{\columnwidth}{!}{%
			\begin{tabular}{rrcrc}
				\toprule
				$\nu p$     & DSLSH ($S_8$) & DSLSH CI & PKNN & PKNN/DSLSH \\
				\midrule
				8 & 7.88 (1.00)  & [6.93, 8.20] & 171.43 & 21.76 \\
				16 & 4.46 (1.77) & [4.01, 4.79] &  85.72 & 19.21 \\
				24 & 2.42 (3.25)  & [2.19, 2.74] &  57.14 & 23.59 \\
				32 & 2,02 (3.89)  & [1.78, 2.20] &  42.86 & 21.17 \\
				40 & 1.53 (5.13)  & [1.33, 1.68] &  34.29 & 22.35 \\
				\bottomrule
			\end{tabular}
		} 
		\caption{\small Strong scaling on \texttt{AHE-5l-5c}, tolerated MCC loss: $10\%$. $S_8$ denotes speedup to single-node ($\nu p = 8$).}
		\label{fig:strong55}
	\end{minipage}
\end{table}


\section{Conclusions and Future Work}

We presented a distributed system for Stratified Locality Sensitive Hashing (DSLSH), allowing for sublinear K-NN prediction times on large physiological time-series repositories in the cloud. We extended the existing LSH AHE prediction analysis by applying the algorithm on datasets in the order of one million points and confirmed the effectiveness of LSH within this domain. We achieved almost perfect scaling on nodes and, depending on the target prediction quality, speedup factors to parallel exhaustive search ranging from one to two orders of magnitude. 
We intend to integrate other and additional similarity metrics to further increase prediction quality on physiological time series.



\end{document}